\documentclass[11pt,a4paper]{article}
\PassOptionsToPackage{hyphens}{url}\usepackage[hyperref]{naaclhlt2018}
\usepackage{times}
\usepackage[hyphenbreaks]{breakurl}
\usepackage[hyphens]{url}
\usepackage{latexsym}
\usepackage{graphicx}
\usepackage{multirow}

\usepackage[utf8]{inputenc}
\usepackage{xcolor}
\usepackage{epstopdf}
\usepackage{tablefootnote}

\aclfinalcopy

\newcommand{\control}[1]{\textcolor{gray}{#1}}

\newcommand{\quotedtext}[1]{{\textquoteleft {#1}\textquoteright}}

\title{Grounding the Semantics of Part-of-Day Nouns Worldwide using Twitter}

\author{David Vilares \\
  Universidade da Coru\~{n}a \\
  FASTPARSE Lab, LyS Group \\
  Departamento de Computaci\'{o}n \\
  Campus de A Elvi\~{n}a s/n, 15071 \\ A Coru\~{n}a, Spain \\
  {\tt david.vilares@udc.es} \\
  \\\And
  Carlos G\'{o}mez-Rodr\'{i}guez \\
  Universidade da Coru\~{n}a \\
  FASTPARSE Lab, LyS Group \\
  Departamento de Computaci\'{o}n \\
  Campus de A Elvi\~{n}a s/n, 15071 \\ A Coru\~{n}a, Spain \\
  {\tt carlos.gomez@udc.es} \\}

\date{}

\begin{document}
\maketitle
\begin{abstract}
The usage of part-of-day nouns, such as \quotedtext{night}, and their time-specific greetings (\quotedtext{good night}), varies across languages and cultures. We show the possibilities that Twitter offers for studying the semantics of these terms and its variability between countries. We mine a worldwide sample of multilingual tweets with temporal greetings, and study how their frequencies vary in relation with local time. The results provide insights into the semantics of these temporal expressions and the cultural and sociological factors influencing their usage.
\end{abstract}

\section{Introduction}\label{section-introduction}

Human languages are intertwined with their cultures and societies, having evolved together, reflecting them and in turn shaping them \cite{Ottenheimer2013,Dediu2013}. Part-of-day nouns (e.g. \quotedtext{morning} or \quotedtext{night}) are an example of this, as their meaning depends on how each language's speakers organize their daily schedule. For example, while the morning in English-speaking countries is assumed to end at noon, the Spanish term (\quotedtext{mañana}) is understood to span until lunch time, which normally takes place between 13:00 and 15:00 in Spain. It is fair to relate this difference to cultural (lunch being the main meal of the day in Spain, as opposed to countries like the {\sc uk}, and therefore being a milestone in the daily timetable) and sociopolitical factors (the late lunch time being influenced by work schedules and the displacement of the Spanish time zones with respect to solar time). Similar differences have been noted for different pairs of languages \cite{Jakel2003} and for cultures using the same language \cite{Sekyi2008}, based on manual study, field research and interviews with natives. Work on automatically extracting the semantics of part-of-day nouns is scarce, as  classic corpora are not timestamped. \newcite{Reiter2003a,Reiter2003b} overcome it by analyzing weather forecasts and aligning them to timestamped simulations, giving approximate groundings for time-of-day nouns and showing idiolectal variation on the term \quotedtext{evening}, but the work is limited to English. 

The relation between language and sociocultural factors implies that the semantics of part-of-day nouns (e.g. 'end of the morning') cannot be studied in isolation from social habits (e.g. 'typical lunch time'). A relevant study of such habits is done by \newcite{walch2016global}, who develop an app to collect sleep habits from users worldwide. While they do not study the meaning of words, their insights are used for validation.

We propose a new approach to study the semantics of part-of-day nouns by exploiting Twitter and the time-specific greetings (e.g. \quotedtext{good morning}) used in different cultures. By mining tweets with these greetings, we obtain a large, worldwide sample of their usage. Since many tweets come with time and geolocation metadata, we can know the local time and country at which each one was emitted. The main contribution of the paper is to show how it is possible to learn the semantics of these terms in a much more extensive way than previous work, at a global scale, with less effort and allowing statistical testing of differences in usage between terms, countries and languages.

\section{Materials and methods}\label{section-materials-methods}

To ground the semantics of greetings we used 5 terms as seeds: \quotedtext{good morning}, \quotedtext{good afternoon}, \quotedtext{good evening}, \quotedtext{good night} and \quotedtext{hello} (a time-unspecific greeting used for comparison). We translated them to 53 languages and variants using Bing translator.\footnote{We used the \href{https://pypi.python.org/pypi/mstranslator}{mstranslator} \textsc{API} for the Bing translator.} 
We use \textsl{italics} to refer to greetings irrespective of the language. 172,802,620 tweets were collected from Sept. 2 to Dec. 7 2016.

For some languages (e.g. Spanish), there is no differentiation between \quotedtext{good evening} and \quotedtext{good night}, and they both are translated to the same expression. For some others, some expressions cannot be considered equivalent, e.g. \quotedtext{good morning} is translated to \quotedtext{bonjour} in French, which is however commonly used as \quotedtext{hello}, or simply as \quotedtext{good day}.

Text preprocessing is not necessary: we rely on metadata, not on the tweet itself, and only the seed words are needed to categorize tweets within a part of day. To clean up the data, we removed retweets, as they last for hours, biasing the temporal analysis. Duplicate tweets were kept, as similar messages from different days and users (e.g. \quotedtext{good night!}) are needed for the task at hand. Tweets need to be associated with a timestamp and country-level geolocation. Tweets have a creation time, composed of a {\sc utc} time and a {\sc utc} offset that varies depending on the time zone. However, most tweets are not geolocated and we must rely on the data provided by the user. This may be fake or incomplete, e.g. specifying only a village. We used fine-grained databases\footnote{http://download.geonames.org/export/dump/} to do the mapping to the country level location and performed a sanity check, comparing the Twitter offset to the valid set of offsets for that country\footnote{http://timezonedb.com}, to reduce the amount of wrongly geolocated tweets.\footnote{Free geolocation {\sc api}'s have rate limits and their use is unfeasible with a large amount of tweets.} 
Comparing the solar and standard time could provide more insights, but this requires a fine-grained geolocation of the tweets.
We obtained a dataset of 10,523,349 elements, available at \burl{https://github.com/aghie/peoples2018_grounding}: 4,503,077 \textsl{good morning}'s, 599,586 \textsl{good afternoon}'s, 214,231 \textsl{good evening}'s, 880,003 \textsl{good night}'s and 4,359,797 \textsl{hello}'s.\footnote{The dataset does not contain tweets from the first two weeks of October due to logistic issues.} 

\section{Results and validation}\label{section-results}

Given a country, some of the tweets are written in foreign languages for reasons like tourism or immigration. This paper refers to tweets written in official or \emph{de facto} languages, unless otherwise specified. Also, analyzing differences according to criteria such as gender or solar time can be relevant. As determining the impact of all those is a challenge on its own, we focus on the primary research question: \emph{can we learn semantics of the part-of-day nouns from simple analysis of tweets?} To verify data quality, \textsl{good morning} tweets were revised: out of 1\,000 random tweets from the {\sc usa}, 97.9\% were legitimate greetings and among the rest, some reflected somehow that the user just started the day (e.g \quotedtext{Didn't get any good morning {\sc sms}}). We did the same for Spain (98,1\% legitimate), Brazil (97.8\%) and India (99.6\%).

Existing work and dated events are used to ratify the results presented below.

\subsection{Worldwide average greeting times}
\label{subsection-results-worldwide}

\begin{table}[t]
\tabcolsep=0.10cm
\centering
\small{
  \begin{tabular}{|l|cccc|}
  \hline
    \bf Country$_{lang}$ &\bf morning & \bf afternoon & \bf night & \bf \control{hello} \\ \hline \hline

Philippines$_{en}$&	08:02:49&13:39:52&00:13:42&\control{14:27:20}\\
Japan$_{ja}$&08:07:28&15:46:50&01:04:19&*\tablefootnote{ Hello translated to \quotedtext{Konnichiwa}, as good afternoon.}\\
South &	\multirow{2}{*}{08:10:07}&\multirow{2}{*}{14:50:52}&\multirow{2}{*}{22:51:48}&\multirow{2}{*}{\control{13:40:19}}\\
Africa$_{en}$&&&&\\
Germany$_{de}$&	08:16:41&13:15:18&23:29:38&\control{14:35:06}\\
Indonesia$_{in}$&	08:17:18&16:25:11&19:02:09&\control{13:55:00}\\
Netherlands$_{nl}$&	08:25:42&14:28:09&23:44:56&\control{14:10:13}\\
Ecuador$_{es}$&	08:32:54&15:03:22&22:10:59&\control{14:37:10}\\
United &	\multirow{2}{*}{08:33:23}&\multirow{2}{*}{13:26:25}&\multirow{2}{*}{21:06:00}&\multirow{2}{*}{\control{13:33:13}}\\
States$_{en}$&&&&\\
Nigeria$_{en}$&	08:34:37&14:11:49&17:19:19&\control{13:40:23}\\
Venezuela$_{es}$&	08:37:03&15:04:00&21:18:05&\control{14:11:07}\\	
Malaysia$_{en}$&	08:39:17&13:31:41&01:02:33&\control{13:56:49}\\
Chile$_{es}$&	08:39:38&15:06:52&00:10:43&\control{14:11:56}\\
Colombia$_{es}$&	08:40:19&15:13:16&21:10:57&\control{14:42:58}\\
Canada$_{en}$&	08:40:30&13:19:33&21:10:57&\control{13:47:40}\\
Mexico$_{es}$&	08:51:04&15:26:35&21:58:24&\control{14:25:37}\\
India$_{en}$&	08:51:24&13:40:00&00:03:12&\control{14:12:54}\\
United&	\multirow{2}{*}{09:06:33}&\multirow{2}{*}{14:30:45}&\multirow{2}{*}{19:49:17}&\multirow{2}{*}{\control{14:13:03}}\\
Kingdom$_{en}$ &&&&\\
Turkey$_{tr}$&	09:16:40&13:12:23&00:41:08&\control{13:56:42}\\
Australia$_{en}$&	09:17:43&15:15:38&20:33:47&\control{13:48:28}\\
Brazil$_{pt}$&	09:18:20&14:47:51&23:31:34&\control{14:26:07}\\
Pakistan$_{en}$&	09:29:12&13:29:28&01:23:05&\control{13:43:58}\\
Russian&	\multirow{2}{*}{09:36:17}&\multirow{2}{*}{13:44:42}&\multirow{2}{*}{23:51:49}&\multirow{2}{*}{\control{14:14:44}}\\
Federation$_{ru}$&&&&\\
Spain$_{es}$&	09:42:41&16:43:57&00:24:28&\control{14:26:33}\\
Argentina$_{es}$&	09:43:47&16:20:05&00:26:55&\control{14:02:03}\\
Greece$_{el}$&	09:46:11&17:12:35&23:28:56&\control{15:01:05}\\
Kenya$_{en}$&	09:57:39&14:15:33&21:44:26&\control{14:07:03}\\
Portugal$_{pt}$&	10:10:22&15:27:35&23:05:25&\control{14:57:34}\\
France$_{fr}$	&12:37:09&*\tablefootnote{The French term for afternoon (après-midi) is not commonly used as part of a greeting.}&00:41:08&\control{14:41:07}\\

\hline
  \end{tabular}}
  \caption{\label{table-avg-greetings} Average local time for the greetings coming from the countries with most data,  sorted by the average time for the greeting \textsl{good morning}. \textsl{Hello} was used as sanity check.}
\end{table}

 \begin{figure*}[t]
 \begin{center}
    \includegraphics[width=1\textwidth,clip]{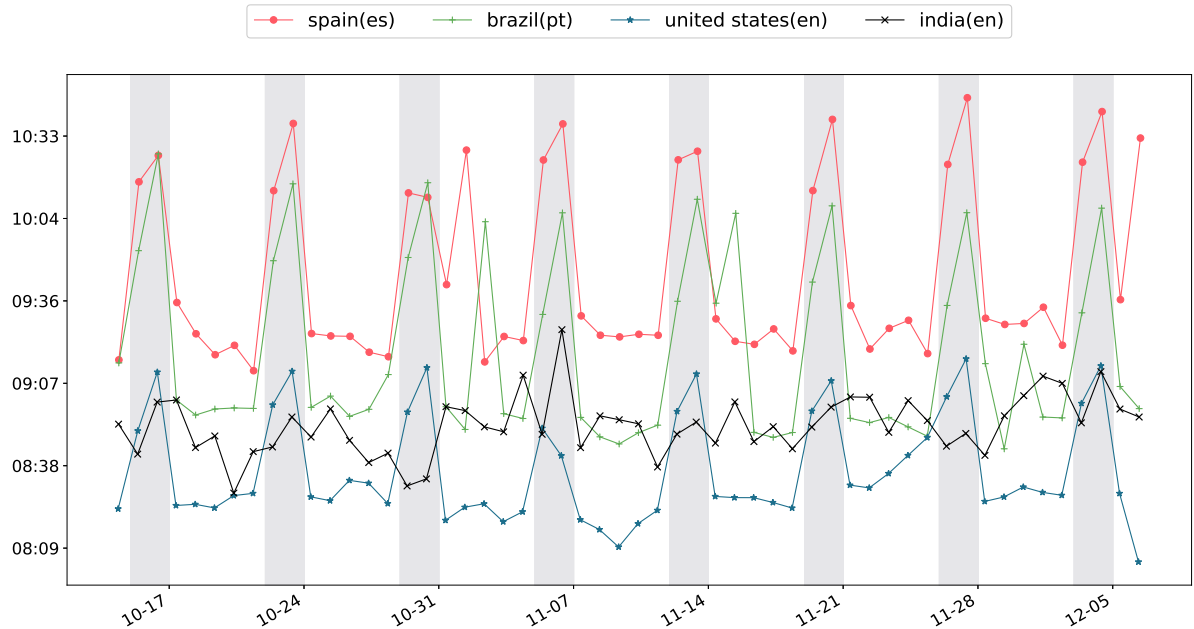}
   \caption{ Average day time for the greeting \textsl{good morning} in different countries ({\sc usa}, Brazil, Spain and India) for a period from mid October to early December, 2016. Weekends are shaded in gray.}
   \label{figure-avg-time-polling-period}
   \end{center}
 \end{figure*}

Table \ref{table-avg-greetings} shows the average greeting times for the countries from which we collected more data. Asian, African and American countries tend to begin the day earlier than Europe (with exceptions, e.g. Germany). The table reflects that countries in southern Europe (e.g. Spain, Portugal or Greece) start the day later than the northern ones (the Netherlands or {\sc uk}). For some countries, e.g. France, this information is known to be biased, as \textsl{good morning} (\quotedtext{bonjour}) is used all along the day. A validation at a fine-grained scale is unfeasible, but the results at the country level are in line with Figure \,3 of \newcite{walch2016global}, e.g., they state that Japan, the {\sc usa} or Germany have earlier wake up times than Spain, Brazil or Turkey.

The average greeting times for \textsl{good afternoon} reveal insights that may stem from cultural differences (e.g. lunch break time). Anglo-Saxon and South Asian countries have the earliest afternoon (with averages between 13:00 and 14:00), while in Mediterranean countries the morning lasts longer (average greeting times for \textsl{good afternoon} around 15:00 or 16:00). A number of countries under the influence of the United Kingdom, such as the United States, Pakistan or India show earlier \textsl{afternoon} times. The opposite happens in South America, historically influenced by Portuguese and Spanish colonialism during the Early modern period, which exhibits later \textsl{afternoon} times. 

This poses interesting questions for future work, such as whether there is a particular reason that could justify this behavior, like having more similar cuisine practices. In this context, the adoption of food practices in colonialism has been already studied by anthropologists and historians \cite{earle2010}. \newcite{trigg2004food} points out how in the early period of the Spanish colonialism in the Americas, they `civilized' the Indigenous community by making them adopt manners, dress and customs. She points that the role of food was specially relevant due to its large social component, and was not limited to the way the food was eaten, but also prepared, served and consumed.

Twitter also reflects differences between countries regarding night life. On the one hand, Anglo-Saxon countries wish \textsl{good night} earlier (from 19:49 in the {\sc uk} to 21:10 in Canada) than other societies. On the other hand, southern European countries go to bed later, and some of them even wish a \textsl{good night} after midnight (e.g. Spain). Comparing to \citet{walch2016global}, we find similar tendencies. For example, in their study Spain, Turkey or Brazil use the smartphone until later than Canada, the {\sc usa} or the {\sc uk}, and therefore they go later to bed. Our Twitter approach also captures the particular case of Japanese mentioned by \citeauthor{walch2016global}: they wake up very early, but use the smartphone until late in the night, suggesting a later bed time.

A fine-grained analysis shows how Twitter captures other cultural and working differences. Figure\,\ref{figure-avg-time-polling-period} charts the average day time for \textsl{good morning} for the {\sc usa}, Brazil, Spain and India during part of the polling period. The time peaks in the weekends for many of the countries, showing that Twitter captures how business and work are reduced during holidays, resulting in later wake up times. 

However, this is not visible in some countries where working conditions are sometimes questioned \cite{MoseIndiaLabour2002}: for India the weekend peak is less pronounced, which can be considered as an indicator that a significant part of its population does not enjoy work-free weekends. 

\begin{figure}[t]
  \begin{center}
    \includegraphics[width=1.0\columnwidth,clip]{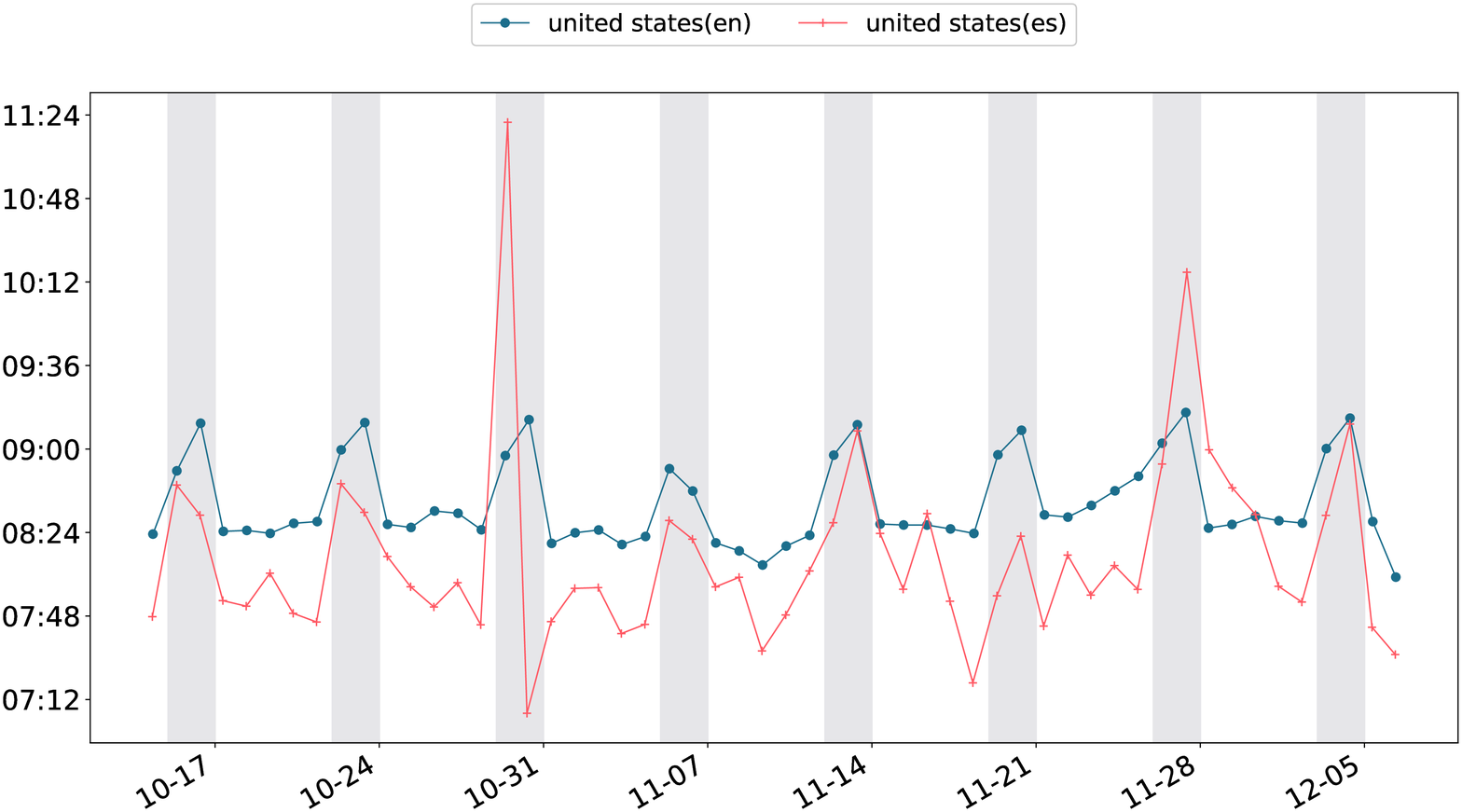} 
    \caption{Average day time for the greeting \quotedtext{good morning} and its Spanish form in the {\sc usa}.}
    \label{figure-avg_morning_us_en_es}
    \end{center}
  \end{figure}

The usage of part-of-day expressions can be helpful to understand more complex issues, such as how foreigners integrate into a country and adapt to its daily schedule. We take the {\sc usa} as example, as it has a large foreign community of Spanish speakers, mainly from Mexico (and in a smaller proportion from other Latin American countries). If we calculate the average day time for the Spanish form of \quotedtext{good morning} (\quotedtext{buenos días}) in the {\sc usa}, we obtain that the result is 08:09, while the corresponding English greeting's average time is 08:33. This is reinforced by Figure\,\ref{figure-avg_morning_us_en_es}, where \quotedtext{buenos días} average day time is consistently lower than \quotedtext{good morning}.\footnote{The peak occurring on 29th October for the Spanish tweets is due to a case of spam that could not be avoided according to the procedure described in \S \ref{section-materials-methods}.} 
This would be in line to their presence in low-wage jobs that require to wake up earlier, e.g. waiter, cleaning or construction work \cite{Flippen2012,Liu2013}. 

It is worth noting that, assuming that these \quotedtext{buenos días} greetings come from latinos, those in the {\sc usa} wake up even earlier than in their countries of origin (see Table \ref{table-avg-greetings}).

Figure\,\ref{figure-avg-time-polling-period} also shows how national holidays influence societies. For example, Nov.\,2 (Day of the Dead) and Nov.\,15 (Proclamation of the Republic) are holidays in Brazil, producing a peak in that country's graph similar to the behavior in the weekends. Similarly, Nov.\,1 (All Saints' Day) and Dec.\,6 (Constitution Day) are holidays in Spain and similar peaks are observed too. From Figure\,\ref{figure-avg_morning_us_en_es} we can see how Thanksgiving (Nov.\,24 in 2016) reflects a four-day weekend in the {\sc usa}: many businesses allow employees to take this holiday from Thursday, resulting into a gradual and increasing peak that spans until Sunday. This is captured by the English \textsl{good morning}s, but not by the Spanish ones. The day after the {\sc usa} 2016 elections (Nov.\,9), a valley occurs on the \textsl{good morning} time for the States (Figure\,\ref{figure-avg-time-polling-period}). The winner was not known until 03:00, suggesting that the distribution of greetings reflects social behaviors in other special events.

\subsection{Daily analysis}

\begin{figure}[t]
 \begin{center}
    \includegraphics[width=1\columnwidth,clip]{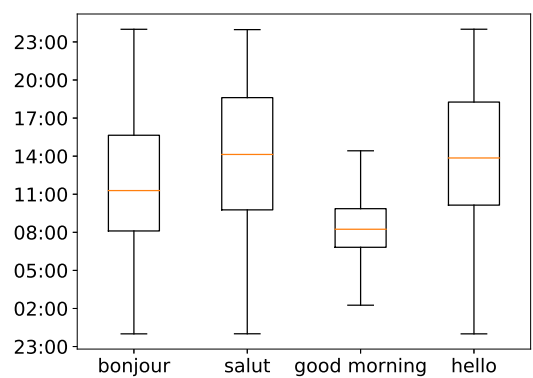}
 \caption{ Box \& whisker plot for the French and English \textsl{good morning}'s and \textsl{hello}'s in Canada.}
  \label{figure-box_canada_en_es}
 \end{center}
\end{figure}

Twitter can be used to do a time-of-day analysis, e.g., as said in \S \ref{subsection-results-worldwide}, \quotedtext{bonjour} is assumed to be used all along the day. To test this, we take Canada, where French and English are official languages. Figure\,\ref{figure-box_canada_en_es} shows how \quotedtext{bonjour} and \quotedtext{salut} (\quotedtext{hello}) are used all along the day, while \quotedtext{good morning} is used in the morning hours. English and French \textsl{hello}'s share a similar distribution.

\begin{figure}[h]
 \begin{center}
    \includegraphics[width=1\columnwidth,clip]{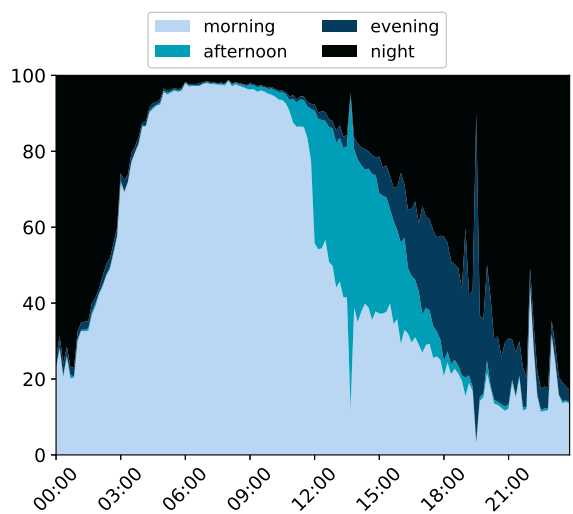}
 \caption{Stacked area chart for the greetings in the {\sc usa}: \% (y axis) vs time (x axis).}
  \label{figure-energy_united_states_en}
 \end{center}
\end{figure}

\begin{figure}[h]
   \begin{center}
     \includegraphics[width=1\columnwidth,clip]{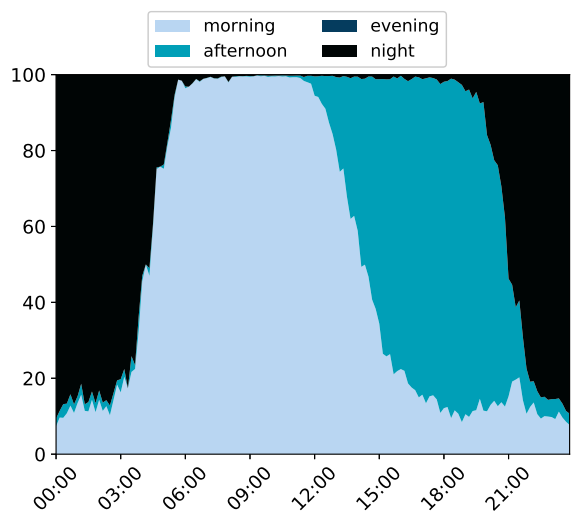}
     \caption{Same as Figure\,\ref{figure-energy_united_states_en}, but for Spain.}
     \label{energy_plot_spain_es}
     \end{center}
\end{figure}

Figure\,\ref{figure-energy_united_states_en} shows a greeting area chart for the {\sc usa}, showing how \quotedtext{good evening} and \quotedtext{good afternoon} are well differentiated, with the transition happening over 16:30. This contrasts to countries such as Spain (Figure\,\ref{energy_plot_spain_es}), where the language has a single word (\quotedtext{tarde}) for \quotedtext{evening} and \quotedtext{afternoon}, whose greeting spans from over 14:00, as the morning ends late (see \S \ref{section-introduction}), to 21:00.

Area plots like these give a clear picture of the semantics of part-of-day nouns, as they depict the exact times when they are used. The precise semantics can be grounded more rigorously using statistical testing to know the exact time intervals at which people significantly use a specific greeting. 

For example, to know when to switch from \textsl{good morning} to \textsl{good afternoon} in Spanish, we can: (1) group the number of \quotedtext{buenos días} (\quotedtext{good morning}) and \quotedtext{buenas tardes} (\quotedtext{good afternoon}) by intervals of 10 minutes, and (2) apply a binomial test to each interval, to determine if one of the greetings is significantly more likely to occur than the other (assuming equal probability of occurrence). For example, for Spain, we obtain that the morning ends at 14:00 (p-value=$2 \times 10^{-8}$ at 14:00, 0.09 at 14:10) and the afternoon starts at 14:40 (p-value becomes statistically significant again with $4 \times 10^{-7}$, showing a significant majority of \textsl{good afternoon}). 

\section{Conclusion}

We crawled Twitter to study the semantics of part-of-day nouns in different countries and societies, showed examples from the polled period and ratified them against existing research and dated events. For space reasons we cannot show insights for all scenarios, but full results are at \burl{https://github.com/aghie/peoples2018_grounding}. 

\section*{Acknowledgments}

DV and CGR receive funding from the European
Research Council (ERC), under the European
Union's Horizon 2020 research and innovation
programme (FASTPARSE, grant agreement No
714150), from the TELEPARES-UDC project
(FFI2014-51978-C2-2-R) and the ANSWER-ASAP project (TIN2017-85160-C2-1-R) from MINECO, and from Xunta de Galicia (ED431B 2017/01).

\bibliographystyle{acl_natbib}
\bibliography{acl2017}

\end{document}